\documentclass[preprint, prX]{revtex4}

\usepackage{amsmath}    
\usepackage{graphicx}   
\usepackage{verbatim}   
\usepackage{color}      
\usepackage{subfigure}  
\usepackage{hyperref}   
\usepackage{amssymb}    
\usepackage{epsfig}
\usepackage{graphics,graphicx}
\usepackage{setspace}
\usepackage{url}
\usepackage{algorithm,algorithmic}
\usepackage{MnSymbol}
\usepackage{multirow}

\begin{document}

\begin{abstract}
Swarm intelligence is a very powerful technique to be used for optimization purposes. In this paper
we present a new swarm intelligence algorithm, based on the bat algorithm. The Bat algorithm is hybridized with differential evolution strategies. Besides showing very promising results of the standard benchmark functions, this hybridization also significantly improves the original bat algorithm.

\textit{To cite paper as follows: I. Jr. Fister, D. Fister, X.-S Yang. 
A hybrid bat algorithm. Elektrotehni\v{s}ki vestnik, 2013, in press.
}

\end{abstract}

\title{A Hybrid Bat Algorithm}

\author{Iztok Fister Jr.}
\altaffiliation{University of Maribor, Faculty of electrical engineering and computer science
Smetanova 17, 2000 Maribor}
\email{iztok.fister@guest.arnes.si}

\author{Du\v{s}an Fister}
\altaffiliation{University of Maribor, Faculty of mechanical engineering
Smetanova 17, 2000 Maribor}
\email{dusan.fister@uni-mb.si}

\author{Xin-She Yang}
\altaffiliation{Middlesex University, School of Science and Technology
London NW4 4BT, UK}
\email{x.yang@mdx.ac.uk}

\maketitle

\section{Introduction}

Nature has always been an inspiration for researchers. In the past, many new nature-inspired
algorithms have been developed to solve hard problems in optimization.
In general, there are two main concepts developed in bio-inspired computation:
\begin{enumerate}
\item evolutionary algorithms,
\item swarm intelligence algorithms.
\end{enumerate}


The Evolutionary algorithms are optimization techniques~\cite{Eiben:2003} that base on the Darwin's principle of survival of the fittest~\cite{Darwin:1859}. It states that in nature the fittest indivi-\\duals have greater chances to survive. The Evolutionary algorithms consist of the following disciplines: genetic algorithms, evolution strategies, genetic programming, evolutionary programming, differential evolution.

Although all these algorithms or methods have been developed independently, they share similar characteristics (like variation operators, selection operators) when solving problems. In fact, the evolutionary algorithms are distinguished by their representation of solutions. For example, the genetic algorithms~\cite{Goldberg:1989,Holland:1992} support the binary representation of solutions, evolution strategies~\cite{baeck:1996,FisterCAI2012} and differential evolution~\cite{storn1997differential,brest2006self,das2011differential} work on real-valued solutions, genetic programming~\cite{Koza:1994} acts on programs in Lisp, while the evolutionary programming~\cite{fogel:1966} behaves with the finish state automata. The Evolutionary algorithms have been applied to a wide range of areas of optimization, modeling and simulation. Essentially, the differential evolution has successfully \\ been employed in the following areas of optimization: function optimization~\cite{shi2005cooperative}, large-scale global optimization~\cite{brest2010large}, graph coloring~\cite{fister2011using}, chemical-process optimization~\cite{babu2006modified}.

Swarm intelligence is a collective behaviour of decentralized, self-organized systems, either natural or artificial. Swarm intelligence was introduced by Beni in 1989. A lot of algorithms have been proposed since then. The Swarm-intelligence algorithms have been applied on continuous as well as combinatorial optimization problems~\cite{parpinelli2011new}. The most well-known classes of the swarm-intelligence algorithms are: particle swarm optimization, ant colony optimization, artificial bee colony, firefly algorithm,
cuckoo search and bat algorithm.

Particle swarm optimization has been successfully applied in problems of the antenna design~\cite{jin2007advances} and electromagnetics~\cite{robinson2004particle}. The Ant colony algorithms have also been used in many areas of optimization~\cite{korovsec2010differential} \cite{parpinelli2002data} \cite{merkle2002ant}. The Artificial bee colony shows good performance in numerical optimization~\cite{karaboga2007powerful} \cite{karaboga2008performance}, in large-scale global optimization~\cite{Fister2012global} and also in combinatorial optimization~\cite{pan2011discrete} \cite{fister2012hybrid} \cite{tasgetiren2010discrete}.

The Cuckoo search algorithm is a very strong method for function optimization and also for engineering optimization problems \cite{yang2010engineering} \cite{yang2009cuckoo}.
The Firefly algorithm shows promising results in function optimization and provides good results also in combinatorial optimization~\cite{fister2012memetic}. 

Echolocation is an important feature of bat behaviour. That means that bats emit a sound pulse and listen to the echo bouncing back from obstacles whilst flying. This phenomenon has inspired Yang~\cite{yang2011bat} to develop the Bat Algorithm (BA). The algorithm shows good results when dealing with lower-dimensional optimization problems, but may become problematic for higher-dimensional problems because of its tending to converge very fast initially. On the other hand, the differential evolution~\cite{Neri:2009} is a typical evolutionary algorithm with differential mutation, crossover and selection that is being successfully applied to continuous function optimization.

To improve the bat-algorithm behaviour for higher-dimensional problems, we hybridized the original bat algorithm by using differential-evolution strategies. This Hybridized Bat Algorithm (HBA) was tested on a standard set of benchmark functions taken from literature. Results of our numerical experiments show that the proposed HBA can significantly improve the performance of the original BA, which can be very useful for the future.
					

The structure of the paper is as follows. In Section 2, we introduced the original BA and the differential evolution algorithm and explain some biological foundations of the bat behaviour. In Section 3 we describe our novel approach to hybridizing the bat algorithm with differential evolution strategies. In Section 4 we illustrate our experiments and discuss the obtained results. The paper ends by presenting our plans for future work on HBA development.

\section{Bat algorithm}
The Bat algorithm was developed by Xin-She Yang in 2010~\cite{Bat:2010}. The algorithm exploits the so-called echolocation of the bats. The bats use sonar echoes to detect and avoid obstacles. It is generally known that sound pulses are transformed into a frequency which reflects from obstacles. The bats navigate by using the time delay from emission to reflection. They typically emit short, loud sound impulses. The pulse rate is usually defined as 10 to 20 times per second. After hitting and reflecting, the bats transform their own pulse into useful information to gauge how far away the prey is. 
The bats are using wavelengths that vary in the range from 0.7 to 17 mm or inbound frequencies of 20-500 kHz. To implement the algorithm, the pulse frequency and the rate have to be defined. The pulse rate can be simply determined in the range from 0 to 1, where 0 means that there is no emission and 1 means that the bats' emitting is their maximum~\cite{gandomi2012bat,tsai2012bat,yang2011review}.

The bat behaviour can be used to formulate a new BA. Yang~\cite{Bat:2010} used three generalized rules when implementing the bat algorithms:
\begin{enumerate}
\item All the bats use an echolocation to sense the distance and they also guess the difference between the food/prey and background barriers in a somewhat magical way.
\item When searching for their prey, the bats fly randomly with velocity $v_{i}$ at position $x_{i}$ with fixed frequency $f_{min}$, varying wavelength $\lambda$ and loudness $A_{0}$. They can automatically adjust the wavelength (or frequency) of their emitted pulses and adjust the rate of pulse emission $r \in [0,1]$, depending on the proximity of their target.
\item Although the loudness can vary in many ways, we assume that it varies from a large (positive) $A_{0}$ to a minimum constant value $A_{min}$.
\end{enumerate}

\begin{algorithm}[H]
\caption{Original Bat Algorithm}
\label{bat}
\small
\begin{algorithmic}[1]
\STATE Objective function $f(x)$, $x=(x_1, ... , x_d )^T$
\STATE Initialize the bat population $x_i$ and $v_i$ for $i = 1 \ldots n$ 
\STATE Define pulse frequency $Q_i \in [Q_{min},Q_{max}]$ 
\STATE Initialize pulse rates $r_i$ and the loudness $A_i$
\STATE while ($t < T_{max}$)  // number of iterations
\STATE \ \ Generate new solutions by adjusting frequency and
\STATE \ \ update velocities and locations/solutions [Eq.(2) to (4)]
\STATE \ \ if($rand(0,1) > r_i$ )
\STATE \ \ \ \ Select a solution among the best solutions
\STATE \ \ \ \ Generate a local solution around the best solution
\STATE \ \ end if
\STATE \ \ Generate a new solution by flying randomly
\STATE \ \ if($rand(0,1) < A_i\ and\ f(x_i) < f(x)$)
\STATE \ \ \ \ Accept the new solutions
\STATE \ \ \ \ Increase $r_i$ and reduce $A_i$
\STATE \ \ end if
\STATE \ \ Rank the bats and find the current best 
\STATE end
\STATE Postprocess results and visualization
\end{algorithmic}
\normalsize
\end{algorithm}

The original BA is illustrated in Algorithm~\ref{bat}. In this algorithm, the bat behaviour is captured into the fitness function of the problem to be solved. It consists of the following components:
\begin{itemize}
\item initialization (lines 2-4),
\item generation of new solutions (lines 6-7),
\item local search (lines 8-11),
\item generation of a new solution by flying randomly (lines 12-16) and
\item find the current best solution.
\end{itemize}

Initialization of the bat population is performed randomly. Generating new solutions is performed by moving virtual bats according to the following equations:
\begin{equation}
\label{velocity}
\begin{aligned}
Q^{(t)}_{i}&=Q_{min}+(Q_{max}-Q_{min}) U(0,1),\\
\mathbf{v}^{(t+1)}_{i}&=\mathbf{v}^{t}_{i}+(\mathbf{x}^{t}_{i}-\mathbf{best}) Q^{(t)}_{i},\\
\mathbf{x}^{(t+1)}_{i}&=\mathbf{x}^{(t)}_{i}+\mathbf{v}^{(t)}_{i},
\end{aligned}
\end{equation}

\noindent where $U(0,1)$ is a uniform distribution. A random walk with direct exploitation is used for the local search that modifies the current best solution according to equation:
\begin{equation}
\label{LS}
\mathbf{x^{(t)}}=\mathbf{best}+\epsilon A^{(t)}_{i}(2 U(0,1)-1),
\end{equation}

\noindent where $\epsilon$ is the scaling factor, and $A^{(t)}_{i}$ the loudness. The local search is launched with the proximity depending on pulse rate $r_i$. The term in line 13 is similar to the simulated annealing behavior, where the new solution is accepted with some proximity depending on parameter $A_i$. In line with this, the rate of pulse emission $r_i$ increases and the loudness $A_i$ decreases. Both characteristics imitate the natural bats, where the rate of the pulse emission increases and the loudness decreases when a bat finds its prey. Mathematically, these characteristics are captured with the following equations:
\begin{equation}
A^{(t+1)}_{i}=\alpha A^{(t)}_{i},\ \ \ \ r^{(t)}_{i} = r^{(0)}_{i}[1-exp(-\gamma \epsilon)],
\end{equation}
 
\noindent where $\alpha$ and $\gamma$ are constants. Actually, $\alpha$ 
\\parameter plays a similar role as the cooling factor in the simulated annealing algorithm that controls the convergence rate of this algorithm. 

\section{Differential evolution}
Differential evolution (DE)\cite{storn1997differential}\cite{das2011differential} is a technique for the optimization introduced by Storn and Price in 1995. 
DE optimizes a problem by maintaining a population of candidate solutions and creates new candidate solutions by combining the existing ones according to its simple formulae, and then keeping whichever candidate solution has the best score or fitness on the optimization problem at hand. DE supports a differential mutation, a differential crossover and a differential selection. In particular, the differential mutation randomly selects two solutions and adds a scaled difference between these to the third solution. This mutation can be expressed as follows:

\begin{equation}
\label{eq:de_mut}
 u_{i}^{(t)}=w_{r0}^{(t)}+F \cdotp (w_{r1}^{(t)}-w_{r2}^{(t)}),\ \ \ \textnormal{for}\ i=1 \ldots NP,
\end{equation}

\noindent where $F \in [0.1,1.0]$ denotes the scaling factor as a positive real number
that scales the rate of modification while $r0,\ r1,\ r2$ are randomly selected vectors in the interval $1 \ldots NP$.

A uniform crossover is employed as a differential crossover by DE. The trial vector is built out of parameter values copied from two different solutions.  Mathematically, this crossover can be expressed as follows:

\begin{equation}
\label{eq:de_xover}
 z_{i,j}=\begin{cases}
          u_{i,j}^{(t)} & \textnormal{rand}_{j}(0,1) \leq \mathit{CR} \vee j=j_{rand}, \\
		  w_{i,j}^{(t)} & \text{otherwise} ,
        \end{cases}
\end{equation}

\noindent where $\mathit{CR} \in [0.0,1.0]$ controls the fraction of parameters that are copied to the trial solution. Note, the relation $j=j_{rand}$ assures that the trial vector is different from the original solution $Y^{(t)}$.

Mathematically, differential selection can be expressed as follows:

\begin{equation}
\label{eq:de_sel}
 w_{i}^{(t+1)}=\begin{cases}
          z_{i}^{(t)} &\text{if } f(Z^{(t)}) \leq f(Y_{i}^{(t)}), \\
		  w_{i}^{(t)} &\text{otherwise}\,.
        \end{cases}
\end{equation}

In the technical sense, the crossover and mutation can be performed in many ways in DE. Therefore, a specific notation was used to describe the variety of these methods (also strategies) in general. For example, "DE/rand/1/bin" denotes that the base vector is randomly selected, one vector difference is added to it, and the number of modified parameters in the mutation vector follows the binomial distribution.

\section{Hybrid Bat Algorithm}
As mentioned above, in this article we propose a new BA, called Hybrid Bat Algorithm (HBA). It was obtained by hybridizing the original BA using the DE strategies. The HBA pseudo-code is illustrated in Algorithm~\ref{hbat}.

\begin{algorithm}[H]
\caption{Hybrid Bat Algorithm}
\label{hbat}
\small
\begin{algorithmic}[1]
\STATE Objective function $f(x)$, $x=(x_1, ... , x_d )^T$
\STATE Initialize the bat population $x_i$ and $v_i$ for $i = 1 \ldots n$ 
\STATE Define pulse frequency $Q_i \in [Q_{min},Q_{max}]$ 
\STATE Initialize pulse rates $r_i$ and the loudness $A_i$
\STATE while ($t < T_{max}$)  // number of iterations
\STATE \ \ Generate new solutions by adjusting frequency and
\STATE \ \ updating velocities and locations/solutions [Eq.(2) to (4)]
\STATE \ \ if($rand(0,1) > r_i$ )
\STATE \ \ \ \ \textbf{Modify the solution using "DE/rand/1/bin"}
\STATE \ \ end if
\STATE \ \ Generate a new solution by flying randomly
\STATE \ \ if($rand(0,1) < A_i\ and\ f(x_i) < f(x)$)
\STATE \ \ \ \ Accept the new solutions
\STATE \ \ \ \ Increase $r_i$ and reduce $A_i$
\STATE \ \ end if
\STATE \ \ Rank the bats and find the current best 
\STATE end
\STATE Postprocess results and visualization
\end{algorithmic}
\normalsize
\end{algorithm}

As a result, HBA differs from the original BA in line 9, where solution is modified using "DE/rand/1/bin" strategy.

\section{Experiments and results}

The goal of our experiments was to show that HBA significantly improves the results of the original BA. For this purpose, two BAs were implemented according to specifications given in Algorithms~\ref{bat} and \ref{hbat} so that a well-selected set of test functions in the literature are used for optimization benchmarks.

The parameters of the two BAs were the same. The dimension of the problem significantly affects the results \\optimization. To test the impact of the problem dimension on the results, three different sets of dimensions were taken into account, i.e., $D=10$, $D=20$, and $D=30$. The functions with dimension $D=10$ were limited to maximally 1,000 generations, the functions with dimension $D=20$ to twice as much, while the functions with dimension $D=30$ to 3,000. The initial loudness was set to $A_0=0.5$ and so was also the initial pulse rate ($r_0=0.5$). The frequency was taken from interval $Q_i \in [0.0,2.0]$. The algorithms optimized each function 25 times and results were measured according to the best, worst, mean, and medium values in these runs. The standard deviation of the mean values was calculated as well.

\subsection{Test suite}

Our test suite consists of five standard functions taken from literature~\cite{yang2010test}. The functions in this test suite are represented below.

\subsubsection{The Griewangk's function}
The aim of the function is overcoming failures to optimize each variable independently. This function is multimodal, since the number of local optima increases with the dimensionality. After dimensionalities are sufficiently high ($n > 30$), multimodality seems to disappear and the problem turns to be unimodal.
\begin{equation} \mathop{f_1(\vec{x})}=-\prod_{i=1}^{D}\cos\left(\frac{x_i}{\sqrt{i}}\right)+\sum_{i=1}^{D}\frac{x_i^2}{4000}+1,\ \end{equation} 
\noindent where $-600\leq x_i \leq600$. The global minimum of the function is at 0.

\subsubsection{The Rosenbrock's function}
As in case of the Rastrigin's function, the Rosenbrock's function too, has its value 0 at the global minimum. The global optimum is inside the parabolic, narrow-shaped flat valley. Variables are strongly dependent on each other, since it is difficult to converge the global optimum.
\begin{equation}
\mathop{f_2(\vec{x_i})}=\sum_{i=1}^{D-1} 100\, (x_{i+1} - x_i^2)^2 + (x_i-1)^2,
\label{sec:rosenbrckrslt}
\end{equation}
\noindent where $-15.00\leq x_i \leq15.00$.

\subsubsection{Sphere function}
\begin{equation} \mathop{f_3(\vec{x_i})}=\sum_{i=1}^{D}x_{i}^2, 
\label{sphere}
\end{equation}
\noindent where $-15.00\leq x_i \leq15.00$.

\subsubsection{The Rastrigin's function}
Based of the Sphere's function, the Rastrigin's function adds cosine modulation to create many local minima. Because of this feature, the function is multimodal. Its global minimum is at 0. 
\begin{equation}
\mathop{f_4(\vec{x_i})}=n*10 + \sum_{i=1}^{n}(x_i^2-10\cos(2\pi x_i)),
\end{equation} 
\noindent where $-15.00\leq x_i \leq15.00$.

 \subsubsection{The Ackley's function}
The complexity of this function is moderate because of the exponential term that covers its surface with numerous local minima. It is based on the gradient slope. Only the algorithm that uses the gradient steepest descent will be trapped in a local optimum. The search strategy, analyzing a wider area, will be able to cross the valley among the optima and achieve better results.
\begin{equation} 
\small
\begin{aligned}
\mathop{f_{5}(\mathbf{x})}=\sum_{i=1}^{n-1}[20+&e-20e^{-0.2\sqrt{0.5(x_{i+1}^2+x_{i}^2)}}-\\
&e^{0.5(\cos(2\pi x_{i+1})+\cos(2\pi x_i))}],\\
\end{aligned}
\normalsize
\end{equation}


\noindent where $-32.00\leq x_i \leq 32.00$. The global minimum of this function is at 0. 

\subsection{PC configuration}

Configuration of PC used in our experiments is as follows:
\begin{itemize}
\item HP Pavilion g4,
\item processor Intel(R) Core(TM) i5 @ 2.40 GHz,
\item memory 8 GB,
\item implemented in C++.
\end{itemize}

\begin{table*}[!htb]
\begin{center}
\caption{The results of experiments}
\small
\begin{tabular}{ | c | c | l | l | l | l | l | l | }
\hline
Alg. & D & Value & $f_1$ & $f_2$ & $f_3$ & $f_4$ & $f_5$ \\ \hline
\hline
\multirow{15}{*}{BA} & \multirow{5}{*}{10} & Best & 3.29E+01 & 1.07E+04 & 5.33E+01 & 6.07E+01 & 1.37E+01 \\
 &  & Worst & 1.73E+02 & 1.58E+06 & 3.11E+02 & 5.57E+02 & 2.00E+01 \\
 &  & Mean & 8.30E+01 & 5.53E+05 & 1.44E+02 & 2.27E+02 & 1.75E+01 \\
 &  & Median & 3.91E+01 & 4.69E+05 & 6.44E+01 & 1.06E+02 & 1.68E+00 \\
 &  & StDev & 6.94E+01 & 4.71E+05 & 1.48E+02 & 2.17E+02 & 1.73E+01 \\
\cline{2-8}
 & \multirow{5}{*}{20} & Best & 8.77E+01 & 3.41E+02 & 2.24E+02 & 7.28E+01 & 2.15E+02 \\
 &  & Worst & 1.43E+02 & 1.02E+03 & 5.72E+02 & 2.02E+02 & 5.87E+02 \\
 &  & Mean & 1.46E+00 & 6.87E+02 & 3.56E+02 & 1.82E+02 & 3.38E+02 \\
 &  & Median & 1.90E+05 & 3.16E+06 & 1.08E+06 & 9.20E+05 & 7.50E+05 \\
 &  & StDev & 1.64E+01 & 2.00E+01 & 1.85E+01 & 1.21E+00 & 1.80E+01 \\
\cline{2-8}
 & \multirow{5}{*}{30} & Best & 1.58E+02 & 4.95E+02 & 3.29E+02 & 8.74E+01 & 3.39E+02 \\
 &  & Worst & 4.18E+02 & 1.67E+03 & 7.80E+02 & 2.64E+02 & 7.82E+02 \\
 &  & Mean & 1.51E+02 & 1.01E+03 & 5.17E+02 & 2.12E+02 & 4.67E+02 \\
 &  & Median & 4.66E+05 & 6.23E+06 & 2.10E+06 & 1.26E+06 & 2.06E+06 \\
 &  & StDev & 1.52E+01 & 2.00E+01 & 1.79E+01 & 1.25E+00 & 1.76E+01 \\
\hline
\multirow{15}{*}{HBA} & \multirow{5}{*}{10} & Best & 2.25E-09 & 6.34E-02 & 4.83E-09 & 5.12E+00 & 6.31E-04 \\
 &  & Worst & 3.97E-05 & 5.10E+02 & 2.89E-03 & 2.38E+01 & 2.00E+01 \\
 &  & Mean & 3.18E-06 & 6.22E+01 & 1.26E-04 & 1.55E+01 & 1.16E+01 \\
 &  & Median & 8.66E-06 & 1.15E+02 & 5.66E-04 & 4.46E+00 & 9.26E+00 \\
 &  & StDev & 1.14E-07 & 7.73E+00 & 1.66E-07 & 1.69E+01 & 1.78E+01 \\
\cline{2-8}
 & \multirow{5}{*}{20} & Best & 1.01E-07 & 9.73E-03 & 4.83E-04 & 1.89E-03 & 3.70E-05 \\
 &  & Worst & 2.96E+01 & 9.24E+01 & 5.47E+01 & 1.77E+01 & 5.48E+01 \\
 &  & Mean & 8.56E-07 & 1.10E-01 & 5.87E-03 & 2.18E-02 & 3.82E-05 \\
 &  & Median & 3.60E+01 & 1.44E+03 & 2.53E+02 & 3.10E+02 & 1.41E+02 \\
 &  & StDev & 2.17E+00 & 2.00E+01 & 1.60E+01 & 6.18E+00 & 1.95E+01 \\
\cline{2-8}
 & \multirow{5}{*}{30} & Best & 6.38E-06 & 8.28E+00 & 3.37E-01 & 1.62E+00 & 5.43E-04 \\
 &  & Worst & 3.57E+01 & 2.17E+02 & 9.97E+01 & 3.98E+01 & 9.85E+01 \\
 &  & Mean & 6.42E-05 & 6.59E+01 & 3.09E+00 & 1.29E+01 & 2.53E-03 \\
 &  & Median & 5.99E+01 & 4.00E+03 & 7.67E+02 & 1.26E+03 & 2.15E+02 \\
 &  & StDev & 3.12E+00 & 2.00E+01 & 1.72E+01 & 5.03E+00 & 1.94E+01 \\
\hline
\end{tabular}
\normalsize
\vspace{-5mm}
\label{tab:detail}
\end{center}
\end{table*}

\subsection{Experiment results}

The results of our extensive numerical experiments are summarized in Table~\ref{tab:detail}. The table represents results of BA and HBA algorithms (column 1) solving the test suite of five functions (denoted as $f_1,f_2,f_3,f_4,f_5$) with dimensions $D=10,$ $20$  $\mathrm{and}$  $	30$, respectively, according to the best, worst, mean, median, and standard deviation values. 

The results of the HBA algorithm show that this algorithm significantly improves the results of the original BA algorithm according to almost all the measures safe for the standard deviation in some cases (e.g., when using Ackley's function). There was no statistical analysis of the results made because they are evidently better for HBA than for BA. 

\begin{figure}[htb]  
\begin{center}
\includegraphics [scale=0.6]{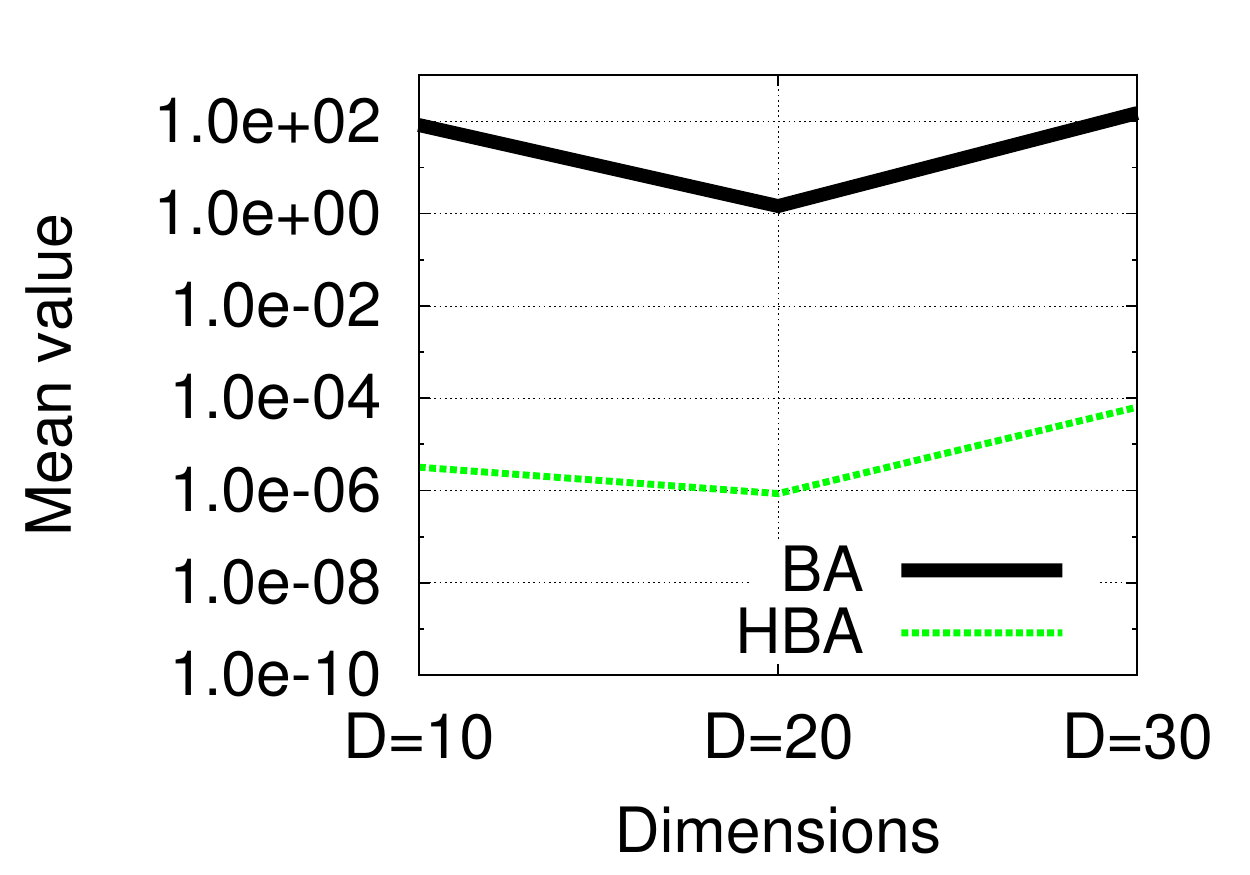}  %
\caption{Mean value of function $f_1$ with various dimensions.}
\label{pic:g1}
\end{center}
\vspace{-5mm}
\end{figure}

\begin{figure}[htb]  
\begin{center}
\includegraphics [scale=0.6]{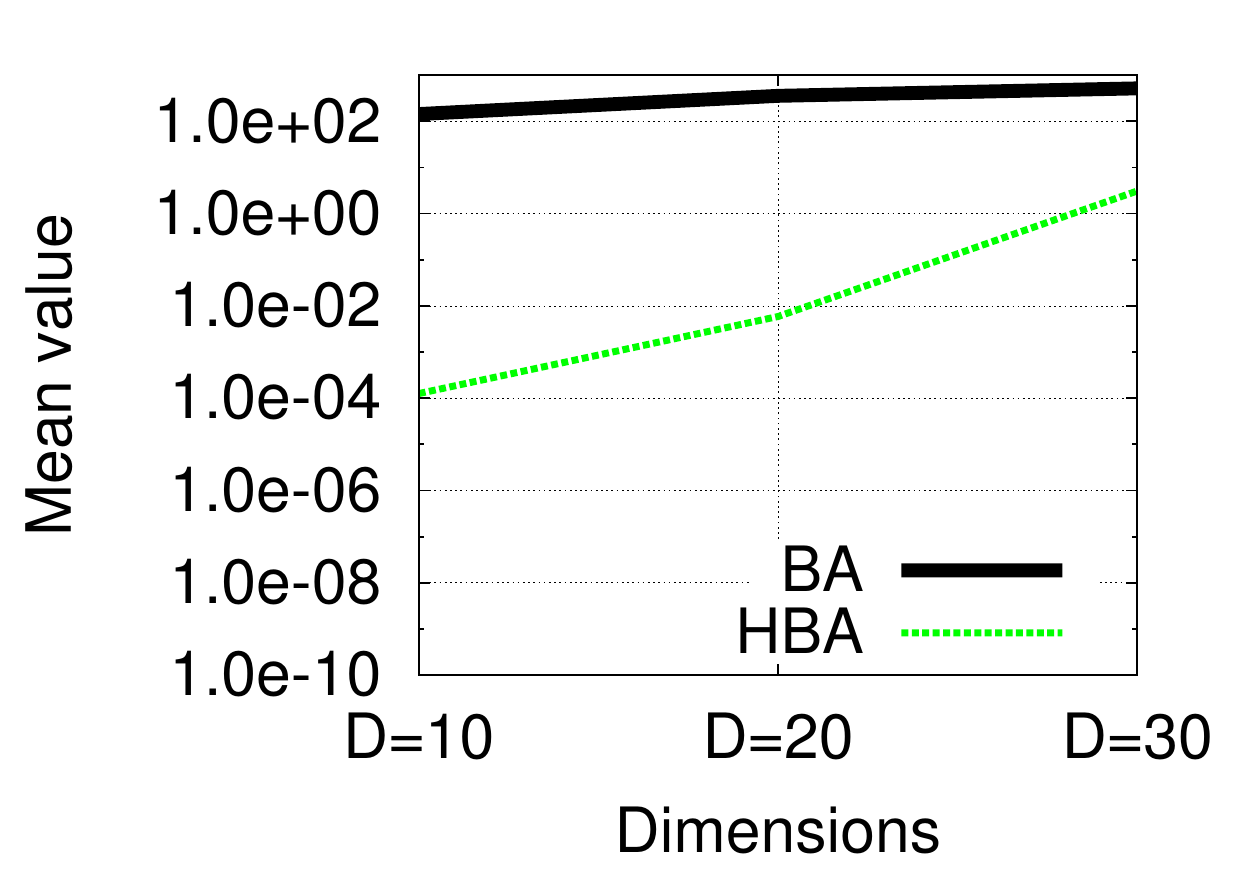}  %
\caption{Mean value of function $f_3$ with various dimensions.}
\label{pic:g2}
\end{center}
\vspace{-5mm}
\end{figure}

\begin{figure}[htb]  
\begin{center}
\includegraphics [scale=0.6]{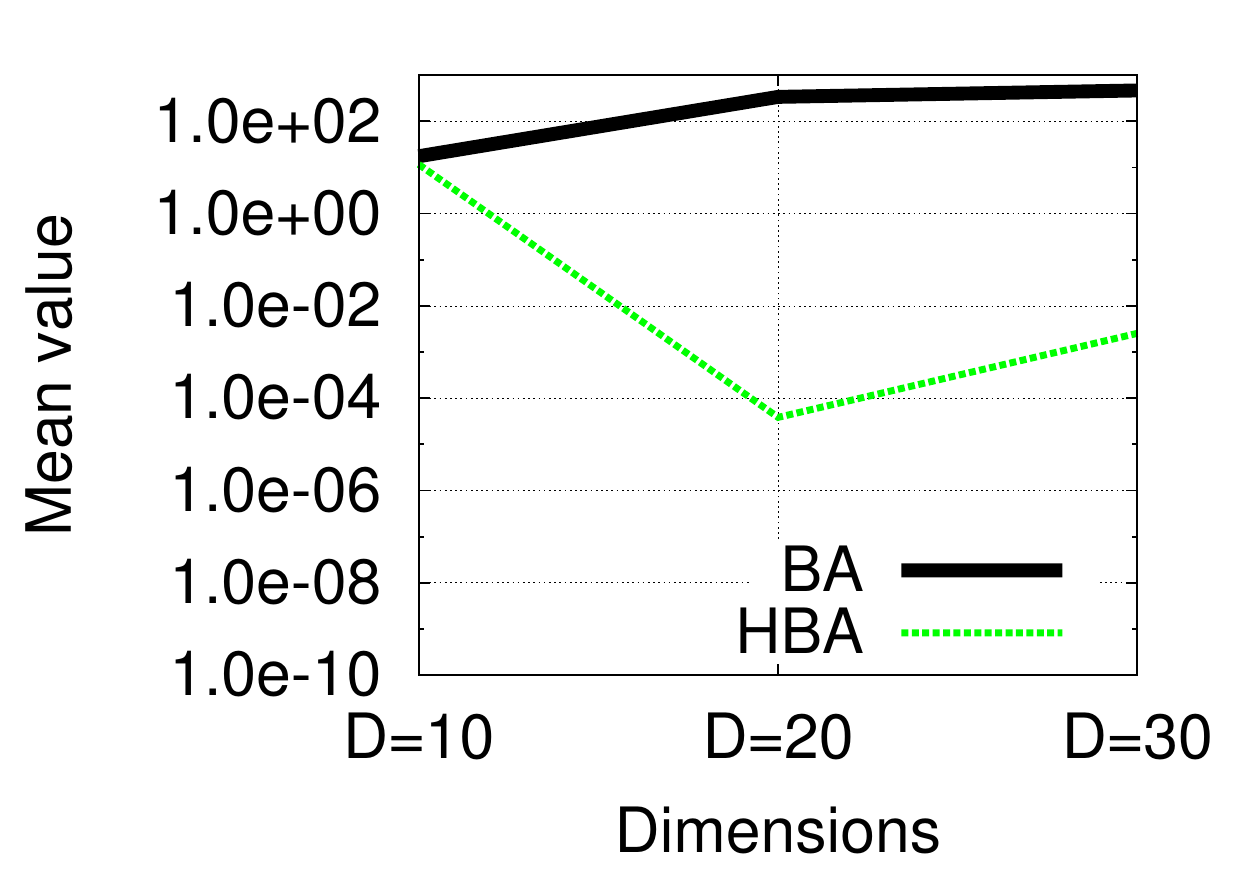}  %
\caption{Mean value of function $f_5$ with various dimensions.}
\label{pic:g3}
\end{center}
\vspace{-5mm}
\end{figure}

To show how the results of the two algorithms (i.e., BA and HBA) vary with the dimensions of the functions, the mean value of functions $f_1$ and $f_3$ with dimensions $D=10$, $D=20$, and $D=30$ are presented in Figs.~\ref{pic:g1}-\ref{pic:g3}. The logarithmic scale is used to display the mean value on the $y$-axis. The higher the mean value, the more difficult the function is to solve.
 
As seen from Fig.~\ref{pic:g1}, the best results are obtained by optimizing function $f_1$ with the dimension $D=20$, while the worst by optimizing the same function with the highest dimension $D=30$. 

It is interesting to note that the mean value of function $f_5$ with dimension $D=10$ is most difficult to obtain for the HBA algorithm, while the same function but with dimension $D=20$ is the easiest to solve. Oppositely, the results of the original BA algorithm show that by increasing the dimensions, the results become worsen.

As seen from Fig.~\ref{pic:g2}, solving function $f_3$ becomes more efficient when the dimensionality of the problem increases. As a result, the most difficult function to solve is function $f_3$ with dimension $D=30$. 

Comparing the results obtained with the BA and the HBA algorithm shows that the results of HBA significantly outperform the results of the original BA algorithm by optimizing the functions $f_1$, $f_3$, and $f_5$. Functions $f_2$ and $f_4$ of the HBA algorithm are also better, but the difference is not significant.

\section{Conclusion}
In this paper we improve the bat algorithm, BA, by developing its new variant, the so-called hybrid bat
algorithm, HBA. HBA is a hybrid of BA with DE strategies. As shown with our experiments, HBA improves significantly the original variant of BA.
In future, HBA will be tested on a large-scale global optimization. Our testing will be extended by using more diverse test function sets and by deepening our parametric study.

\bigskip{\small \smallskip\noindent Updated 30 April 2013.}
\end{document}